\documentclass[10pt,twocolumn,letterpaper]{article}
\PassOptionsToPackage{dvipsnames,table}{xcolor}

\usepackage{cvpr}
\usepackage[dvipsnames]{xcolor}

\newcommand{\method}[1]{RobustVisRAG}
\newcommand{\methodM}[1]{Disentangled Prompt for Robust Generation}

\newcommand{\promptengineM}[1]{Causality Prompting Token}
\newcommand{\promptengine}[1]{CPT}

\newcommand{\promptA}[1]{CLS}
\newcommand{\promptB}[1]{NCP}
\newcommand{\promptBM}[1]{Non-Causal Prompt}

\newcommand{\dataset}[1]{Distortion-VisRAG}

\newcommand{\lossa}[1]{NCDM}
\newcommand{\lossaM}[1]{Non-Causal Distortion Modeling}

\newcommand{\lossb}[1]{CSA}
\newcommand{\lossbM}[1]{Causal Semantic Alignment}

\definecolor{cvprblue}{rgb}{0.21,0.49,0.74}

\usepackage{times}
\usepackage{graphicx}
\usepackage{amsmath}
\usepackage{amssymb}
\usepackage{booktabs}
\usepackage{rotating}
\usepackage{paralist}
\usepackage{caption}
\usepackage{subcaption}
\usepackage[linesnumbered,boxed]{algorithm2e}
\usepackage{arydshln}
\usepackage{array}
\usepackage{multirow}
\usepackage{ragged2e}
\usepackage{booktabs}
\usepackage{arydshln}
\usepackage{makecell}

\usepackage[dvipsnames]{xcolor}
\usepackage[accsupp]{axessibility}
\usepackage{times}
\usepackage{array}
\usepackage{ragged2e}
\usepackage{booktabs}
\usepackage{wrapfig}

\usepackage{graphicx}
\usepackage{amsmath}
\usepackage{amssymb}
\usepackage[pagebackref,breaklinks,colorlinks,citecolor=blue,linkcolor=blue,bookmarks=false]{hyperref}
\usepackage{makecell}
\usepackage[accsupp]{axessibility}
\usepackage{color, colortbl}
\definecolor{LightCyan}{rgb}{0.88,1,1}

\hypersetup{
    colorlinks=true,
    urlcolor=magenta,
    citecolor=blue,
}

\usepackage[T1]{fontenc}
\usepackage{multirow}
\usepackage{booktabs}
\usepackage{comment}
\usepackage{color}
\usepackage{amsmath}

\usepackage{textcomp}
\usepackage{siunitx}
\usepackage{tabulary}
\usepackage{makecell}
\usepackage{multirow}
\usepackage{booktabs}
\usepackage{hyperref}

\usepackage{etoolbox}
\makeatletter

\newcolumntype{P}[1]{>{\centering\arraybackslash}p{#1}}
\newcolumntype{M}[1]{>{\centering\arraybackslash}m{#1}}
\let\ts@includegraphics\includegraphics

\usepackage{float}
\usepackage{lipsum}   
\usepackage{rotating} 
\usepackage{stfloats} 
\usepackage[export]{adjustbox} 


\usepackage[capitalize]{cleveref}
\crefname{section}{Sec.}{Secs.}
\Crefname{section}{Section}{Sections}
\Crefname{table}{Table}{Tables}
\crefname{table}{Tab.}{Tabs.}

\def\paperID{36281} 
\def\confName{CVPR}
\def\confYear{2026}

\begin{document}
\title{RobustVisRAG: Causality-Aware Vision-Based Retrieval-Augmented Generation under Visual Degradations}

\vspace{-20mm}\author{
    I-Hsiang Chen\textsuperscript{1}
    \quad Yu-Wei Liu\textsuperscript{1}
    \quad Tse-Yu Wu\textsuperscript{1} 
    \quad Yu-Chien Chiang\textsuperscript{1} \\
    \quad Jen-Chieh Yang\textsuperscript{1}
    \quad Wei-Ting Chen\textsuperscript{2}
    \\\\
    \hspace{-8mm}\textsuperscript{1}National Taiwan University\quad \textsuperscript{2}Microsoft
}

\maketitle


\begin{abstract}
Vision-based Retrieval-Augmented Generation (VisRAG) leverages vision-language models (VLMs) to jointly retrieve relevant visual documents and generate grounded answers based on multimodal evidence. 
However, existing VisRAG models degrade in performance when visual inputs suffer from distortions such as blur, noise, low light, or shadow, where semantic and degradation factors become entangled within pretrained visual encoders, leading to errors in both retrieval and generation stages. 
To address this limitation, we introduce \method{}, a causality-guided dual-path framework that improves VisRAG robustness while preserving efficiency and zero-shot generalization.
\method{} uses a non-causal path to capture degradation signals through unidirectional attention and a causal path to learn purified semantics guided by these signals.
Together with the proposed Non-Causal Distortion Modeling and Causal Semantic Alignment objectives, the framework enforces a clear separation between semantics and degradations, enabling stable retrieval and generation under challenging visual conditions.
To evaluate robustness under realistic conditions, we introduce the \dataset{} dataset, a large-scale benchmark containing both synthetic and real-world degraded documents across seven domains, with 12 synthetic and 5 real distortion types that comprehensively reflect practical visual degradations.
Experimental results show that \method{} improves retrieval, generation, and end-to-end performance by 7.35\%, 6.35\%, and 12.40\%, respectively, on real-world degradations, while maintaining comparable accuracy on clean inputs. 
Project page: \url{https://robustvisrag.github.io/}
\end{abstract}

\newcommand\blfootnote[1]{%
\begingroup
\renewcommand\thefootnote{}\footnote{#1}%
\addtocounter{footnote}{-1}%
\endgroup
}


\setlength{\parskip}{0em}


\begin{figure}[t!]
    \centering
    \vspace{-5pt}
    \begin{subfigure}{0.48\columnwidth}
        \centering
        \includegraphics[width=\linewidth,
        trim=10pt 10pt 10pt 10pt,
        clip]{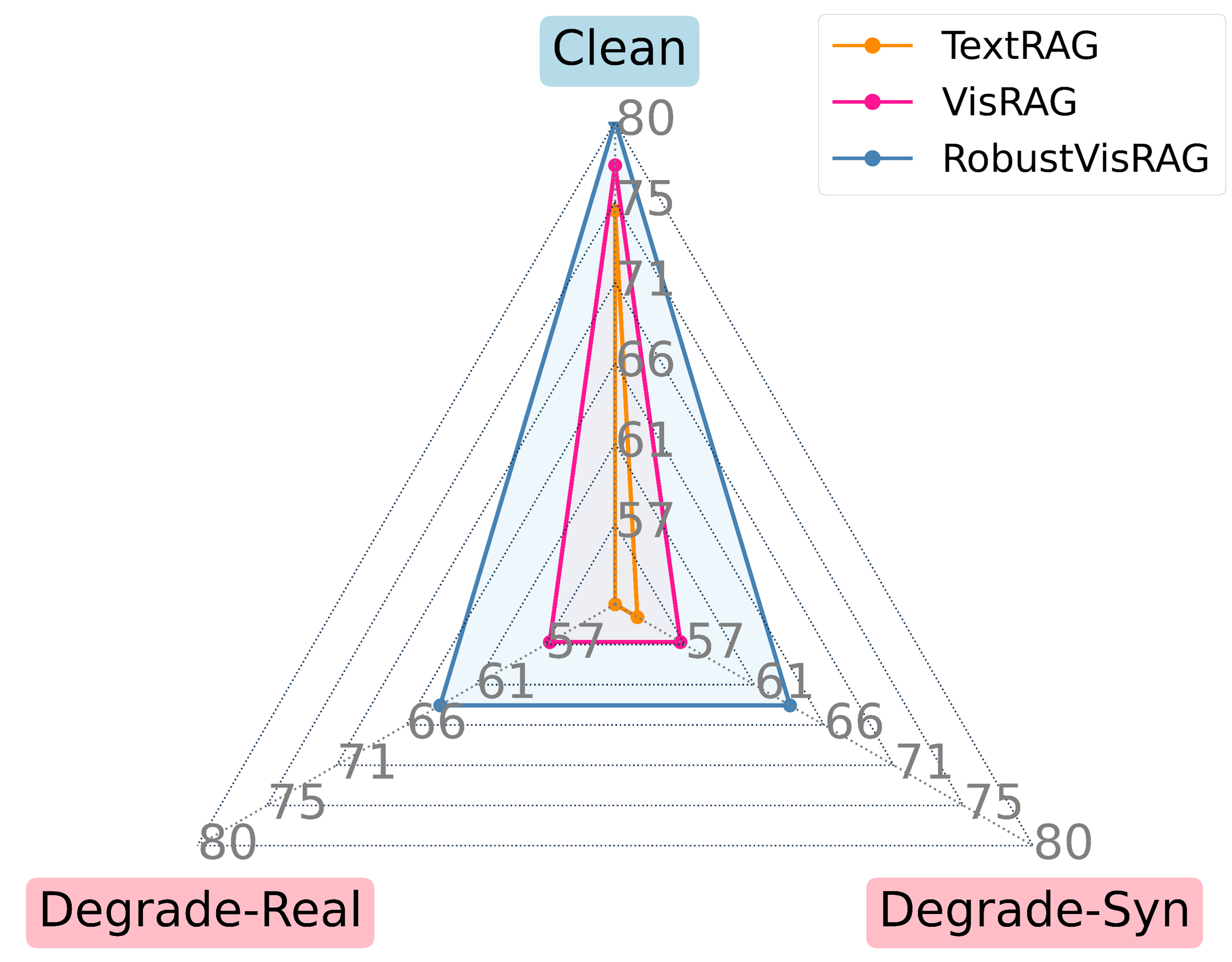}
        \caption{Retrieval}
        \label{fig:teaser_retrieval}
    \end{subfigure}
    \hspace{0.1cm}
    \begin{subfigure}{0.48\columnwidth}
        \centering
        \includegraphics[width=\linewidth,
        trim=10pt 10pt 10pt 10pt,
        clip]{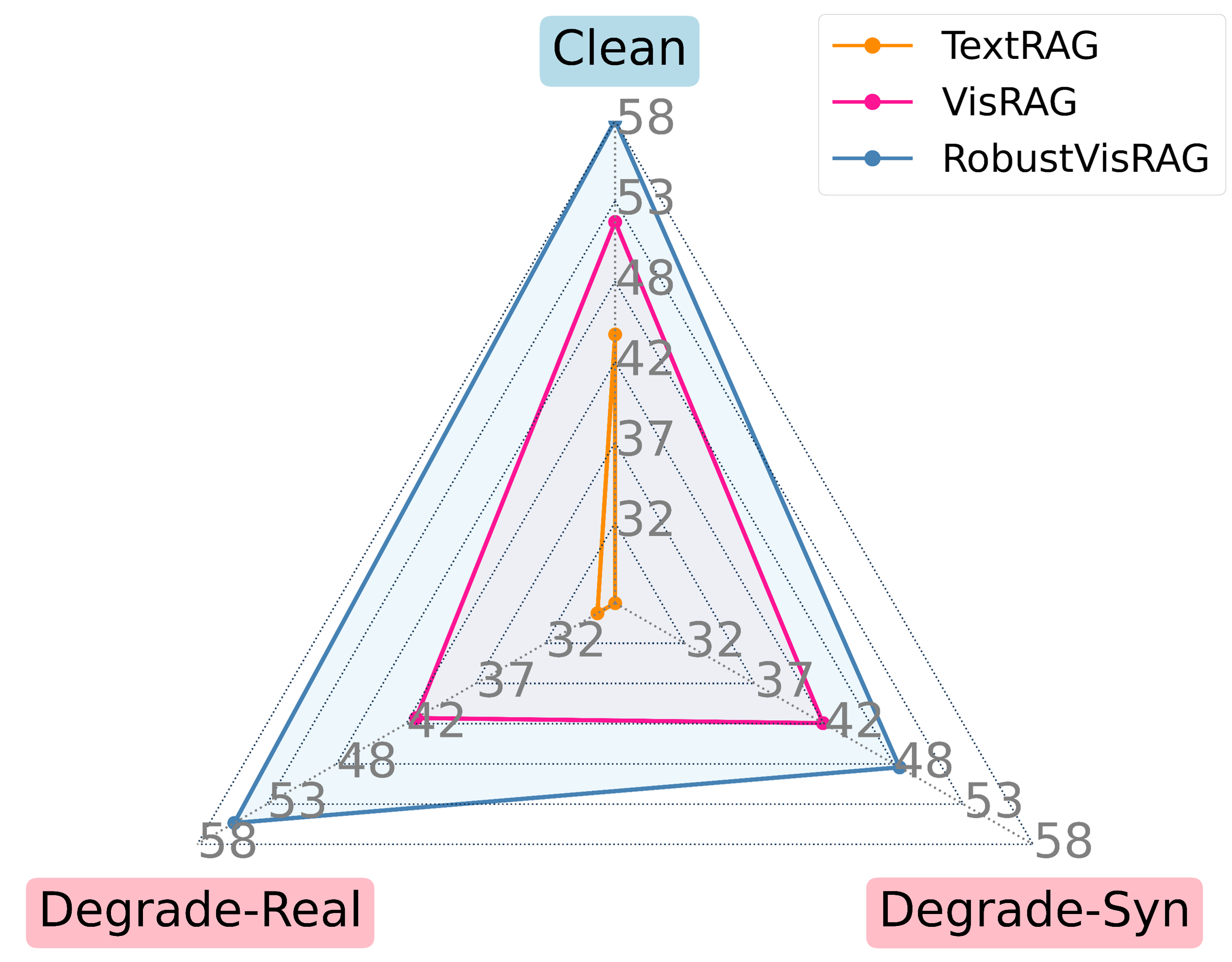}
        \caption{Generation}
        \label{fig:teaser_generation}
    \end{subfigure}

\vspace{12pt}
\begin{subfigure}{0.48\columnwidth}
    \centering
    \includegraphics[
        width=\linewidth,
        trim=5pt 5pt 5pt 5pt, 
        clip
    ]{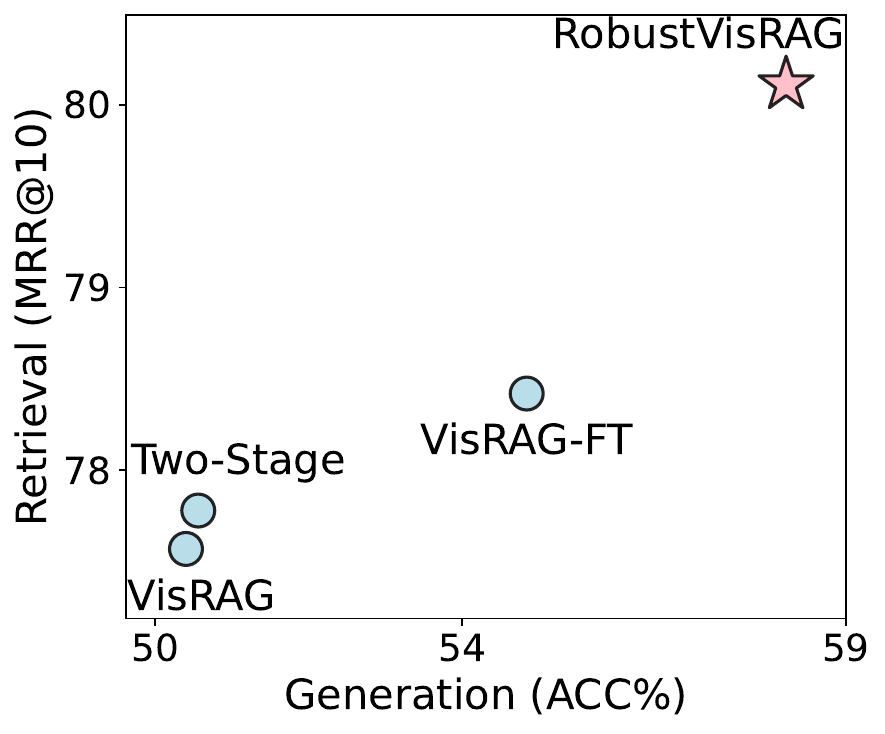}
    \caption{End-to-End (Clean)}
    \label{fig:teaser_overview1}
\end{subfigure}
\hspace{0.01cm}
\begin{subfigure}{0.48\columnwidth}
    \centering
    \includegraphics[
        width=\linewidth,
        trim=5pt 5pt 5pt 5pt,
        clip
    ]{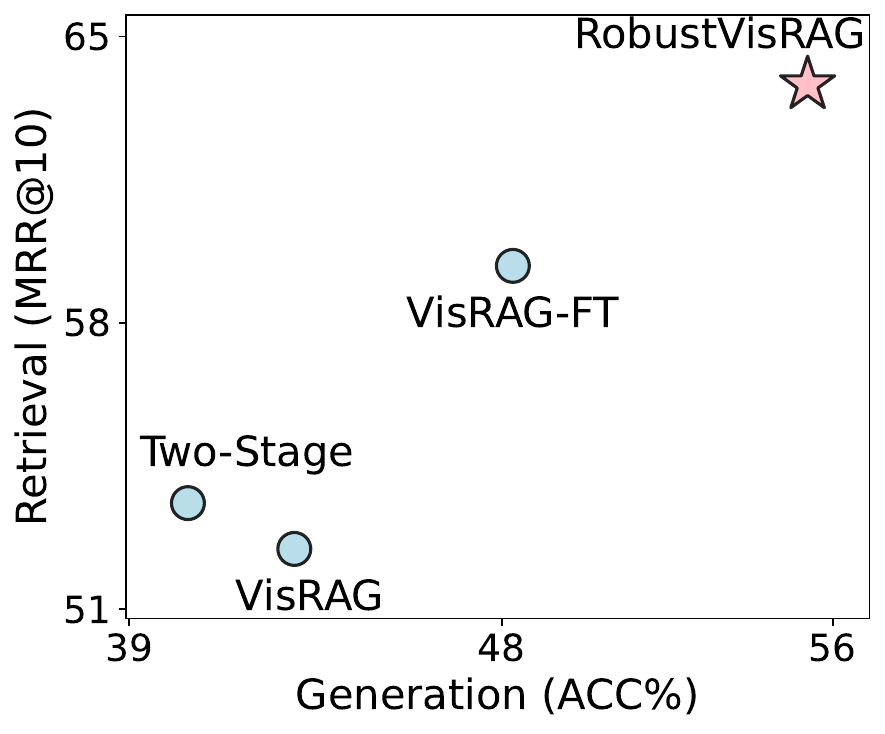}
    \caption{End-to-End (Degrade)}
    \label{fig:teaser_overview2}
\end{subfigure}

    \vspace{-5pt}
\caption{\textbf{Illustration of \method{}'s capabilities.}
(a) Retrieval performance under clean, synthetic degradation (Degrade-Syn) and real-degradation (Degrade-Real) scenarios. 
(b) Generation performance using the retrieved documents from \method{} as input.
(c)(d) End-to-end retrieval–generation performance on clean and degraded data, evaluated under the same baselines including VisRAG, its fine-tuned variants (VisRAG-FT)~\cite{schlarmann2024robust}, and a Two-Stage restoration pipeline~\cite{potlapalli2023promptir}. 
Across all settings, both TextRAG and VisRAG show notable performance drops under degraded inputs, whereas \method{} preserves clean accuracy and significantly improves robustness in degraded conditions.}
    \label{fig:teaser}

\end{figure}

\section{Introduction}
Large language models (LLMs) and vision-language models (VLMs) have shown strong reasoning and generation abilities across diverse tasks~\cite{alayrac2022flamingo, li2022blip,liu2023llava, zhao2023survey, openai2023gpt,Qwen-VL}. 
However, their parametric nature fixes internal knowledge after training, often leading to hallucinations or outdated predictions~\cite{ji2023survey, bang2023multitask}. 
Retrieval-augmented generation (RAG)~\cite{guu2020retrieval, lewis2020retrieval, yu2023augmentation,yu2024visrag} addresses this limitation by incorporating external knowledge during generation, improving factuality and interpretability.
Existing RAG methods can be categorized into Text-based RAG (TextRAG)~\cite{guu2020retrieval} and Vision-based RAG (VisRAG)~\cite{yu2024visrag}. 
TextRAG operates on textual inputs, retrieving relevant passages for generation. 
When document content appears as images, it relies on multi-stage parsing with layout analysis and OCR~\cite{pfitzmann2022doclaynet,denk2023docaid}, which accumulates recognition errors~\cite{yu2024visrag}, discards visual cues, and fails to capture non-textual information such as figures or charts. In contrast, VisRAG leverages VLMs to directly encode visual inputs, avoiding parsing errors and preserving spatial and visual context. 
This design enables effective multimodal reasoning and alignment during generation.

Although both TextRAG and VisRAG reduce hallucinations and outdated predictions by retrieving external evidence, their performance drops when either the query or the corpus contains degraded document images affected by blur, noise, low light, shadow, or compression artifacts~\cite{usama2025analysing}, as shown in \cref{fig:teaser_retrieval} and \cref{fig:teaser_generation}. 
In TextRAG, such degradations cause recognition and parsing failures in OCR and layout detection, resulting in incomplete or erroneous retrieval. 
In VisRAG, visual distortions corrupt the embeddings extracted by VLM encoders, where semantic and distortion factors intertwine, leading to retrieval mismatches and unstable generation. 
Degradations induce a dual failure mode in the VisRAG pipeline: the model may retrieve incorrect evidence due to corrupted visual representations, and even when the correct evidence is retrieved, degraded inputs can still mislead the generation process. 
These challenges highlight the need for robustness analysis and mitigation within the VisRAG pipeline under degraded conditions.

To address the above challenges, an intuitive two-stage strategy is to apply existing image restoration techniques~\cite{son2020urie,wei2023uniir} to improve the visual quality of degraded images and use the enhanced results as inputs to VisRAG. 
However, under the degraded VisRAG setting, such restoration-based two-stage pipelines do not consistently translate perceptual improvements into retrieval or generation gains. Alternatively, fine-tuning the VLM itself offers a more direct solution. %
Parameter-efficient fine-tuning (PEFT)~\cite{hu2022lora} offers a low-cost way to adapt VLMs, but its limited representational capacity hinders effective recovery of degradation-corrupted embeddings. 
In contrast, full fine-tuning (FFT) enhances adaptability to degraded inputs but requires substantial computational resources and often overfits to distortion patterns while forgetting pretrained knowledge~\cite{jiang2024refine}. 
Moreover, both fine-tuning strategies remain fundamentally limited, as they lack explicit causal guidance to disentangle semantic and distortion factors, which is essential for achieving robust VisRAG under visual degradations. 
As shown in~\cref{fig:teaser_overview2}, existing two-stage and fine-tuning strategies provide limited performance gains under degraded conditions, highlighting the persistent challenge of distortion robustness in VisRAG. %

To address the vulnerability of VisRAG systems under visual degradations, we propose \method{}, a causality-guided dual-path framework that explicitly separates semantic and degradation information during visual encoding.
\method{} augments the vision encoder with two complementary pathways:
a non-causal path that aggregates degradation cues using a unidirectional attention mechanism, and
a causal path that focuses on semantic aggregation and learns purified semantics under the guidance of the degradation signals extracted by the non-causal path.
To ensure functional specialization of both pathways, we introduce two learning objectives:
Non-Causal Distortion Modeling (NCDM) to enforce structured degradation representation, and
Causal Semantic Alignment (CSA) to guide semantic purification and prevent degradation leakage into semantic embeddings.
Through joint optimization of these components within a single forward pass, \method{} effectively disentangles semantic and degradation factors, substantially improving retrieval and generation robustness under degraded visual conditions as shown in \cref{fig:teaser}.

To evaluate real-world robustness, 
we construct the \dataset{} dataset, extending VisRAG~\cite{yu2024visrag} 
with large-scale synthetic and real-world degraded document images.  
The dataset additionally includes a real-world subset captured under practical conditions such as low light, shadow, and paper damage, narrowing the gap between simulated and natural degradations.
The dataset contains 367K question–document pairs across seven document understanding domains 
(e.g., scientific papers, charts, forms, slides, and handwritten notes), 
covering 17 degradation types at multiple severity levels.  
This benchmark enables systematic evaluation of retrieval and generation robustness.

The main contributions of this work are summarized as follows:
\begin{compactitem}
\item 
We propose \method{}, a causality-guided dual-path framework that disentangles semantic and degradation factors during visual encoding to improve VisRAG robustness under degraded conditions without additional inference cost.
Extensive experiments show that \method{} generalizes to real-world degradations in our benchmark, improving retrieval, generation, and end-to-end performance by 7.35\%, 6.35\%, and 12.40\%, respectively, while maintaining comparable performance on clean data.

\item We introduce the \dataset{} dataset, a VisRAG-specific benchmark designed for joint evaluation of retrieval, generation, and end-to-end robustness under synthetic and real-world degradations.
\end{compactitem}

\section{Related Work}

\subsection{Retrieval-Augmented Generation}
RAG enhances LLMs by coupling retrieval with generation, grounding answers in external evidence rather than internal parameters~\cite{guu2020retrieval,shi2023replug,yu2023augmentation}.
RAG system includes a retriever for locating relevant information and a generator for producing responses based on the query and retrieved context.

\noindent\textbf{Text-based RAG.}
These methods operate purely on text by applying OCR~\cite{pfitzmann2022doclaynet,denk2023docaid,zhong2019publaynet,jaume2019funsd} to document images before retrieval and generation.
However, OCR errors and layout loss in complex documents (e.g., tables or charts) cause fragmented information and grounding issues~\cite{liu2021dense,xu2024chatqa}, which are further exacerbated under degradations such as blur, noise, or compression~\cite{courtney2021sediqa}. While recent studies~\cite{most2025lost} suggest that OCR-based pipelines can exhibit competitive robustness under certain document quality conditions, their performance remains sensitive to dataset characteristics and OCR quality.

\noindent\textbf{Vision-based RAG.}
To address the limitations of text-based RAG, recent studies extend RAG to the visual domain~\cite{yu2024visrag,faysse2024colpali,ding2023pdfvqa}.
VisRAG replaces textual modules with vision-language retrievers and generators that encode document images and queries in a shared embedding space, preserving layout and visual cues while reducing OCR errors.
However, current models still assume ideal image quality and degrade under real-world conditions, underscoring the need for robustness-oriented VisRAG frameworks.

\subsection{VLM Robustness under Degradation}
A degradation-robust VLM is essential for reliable VisRAG systems, as both retrieval and generation depend on visual representation quality.
Recent studies show that existing VLMs experience performance drops under visual degradations~\cite{usama2025analysing}.
Although no method is explicitly designed for degradation-aware multimodal learning, prior works have explored potential directions to alleviate this issue, including input enhancement and fine-tuning strategies. 
One common approach is the two-stage strategy, which enhances input quality before feeding images into the model.
Image restoration methods~\cite{zamir2021multi, yang2023visual, chen2025unirestore} improve perceptual fidelity but may not preserve semantic consistency~\cite{chen2025exploring, chen2022rvsl, chen2022sjdl}, limiting downstream gains.
Another approach is fine-tuning: parameter-efficient methods~\cite{hu2022lora} offer lightweight adaptability but limited capacity, while full fine-tuning enhances degraded performance but incurs high computational cost and risks catastrophic forgetting~\cite{jiang2024refine}, hindering generalization.
Moreover, methods such as TeCoA~\cite{mao2022understanding} and FARE~\cite{schlarmann2024robust} enhance robustness by adversarially fine-tuning the vision encoders of VLMs, improving their resistance to $\ell_p$-bounded perturbations. 
However, their improvements are confined to small, controllable pixel perturbations and may not generalize to the natural degradations considered here, such as blur, low light, compression, and shadow.
Overall, while robustness of VLMs under degradations has been widely studied, robust vision-language understanding within VisRAG pipelines remains underexplored. 

\subsection{Causality Learning}
Causality learning provides a principled framework for understanding the underlying relationships between data attributes and model predictions~\cite{imbens2015causal,spirtes2000causation,spirtes2000causation}.
In vision research, Structural Causal Models have been used to disentangle true semantic causes from spurious correlations, promoting debiasing and domain generalization~\cite{xu2023multi,wang2022contrastive,lv2022causality}.
Building on this foundation, causal attention~\cite{wang2021causal} and causal prototype learning~\cite{liu2024unbiased} further incorporate causal reasoning into neural architectures, enhancing interpretability and robustness in visual understanding.
Recently, causal reasoning has also been introduced into VLMs to improve compositional reasoning and multimodal alignment~\cite{gao2025capl,parascandolo2025cogt,li2025mucr}.
However, these methods mainly operate at the semantic level and assume clean inputs.
In real-world scenarios, visual signals are often affected by various forms of visual degradation, which act as latent causal factors that directly influence feature representations.
Yet current causal VLM frameworks rarely model this “degradation → representation” pathway, leaving semantic features entangled with degradation cues and reducing robustness under real-world conditions.
\section{Proposed Method}

\begin{figure*}[t!]
    \centering
    \setlength{\tabcolsep}{2pt}

    \begin{subfigure}[t]{0.28\textwidth}
        \centering
        \includegraphics[width=\linewidth, height=7cm, keepaspectratio=false]{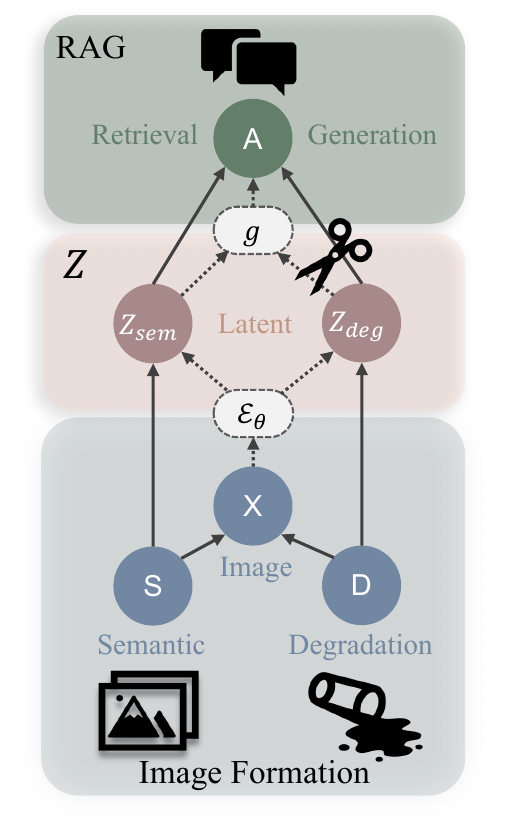}
        \caption{Structural Causal Model}
        \label{fig:scm}
    \end{subfigure}
    \begin{subfigure}[t]{0.30\textwidth}
        \centering
        \includegraphics[width=\linewidth, height=7cm, keepaspectratio=false]{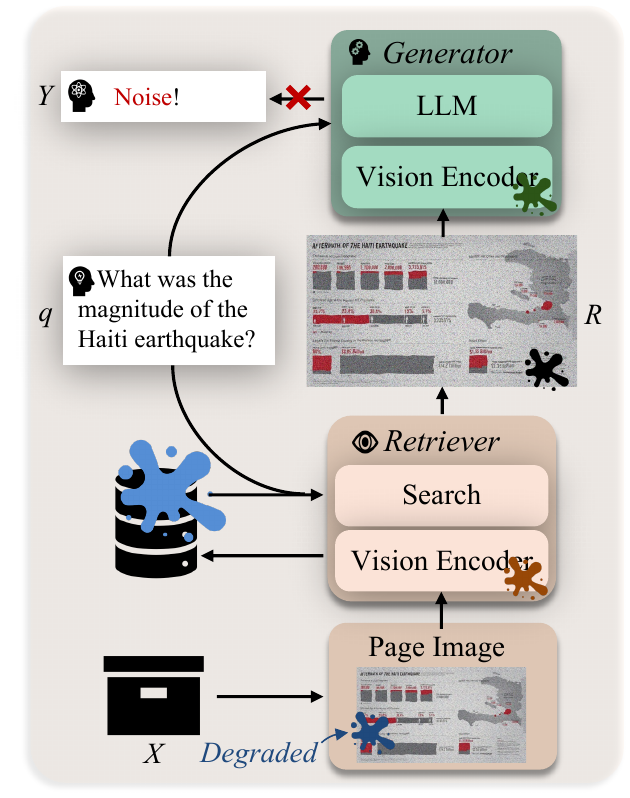}
        \caption{Vision-based RAG}
        \label{fig:rag}
    \end{subfigure}
    \begin{subfigure}[t]{0.30\textwidth}
        \centering
        \includegraphics[height=7cm]{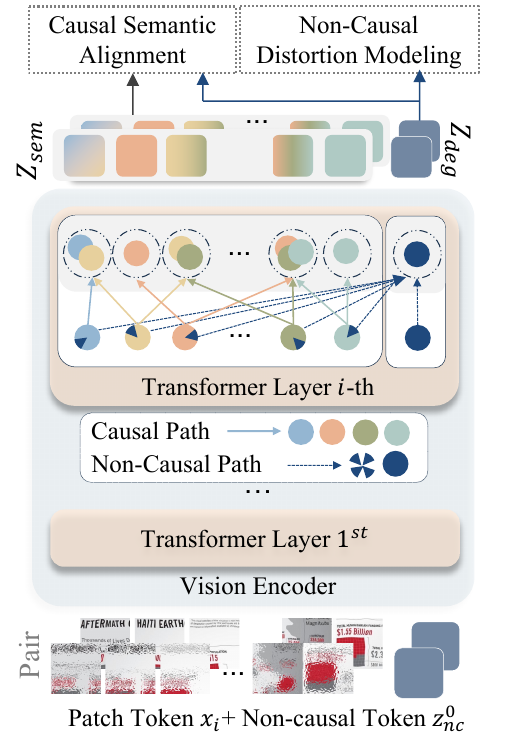}
        \caption{\method{}.}
        \label{fig:robustvisrag}
    \end{subfigure}

\caption{\textbf{Overview of \method{}.} 
(a) Structural causal model of VisRAG under visual degradations. 
(b) Architecture of the vanilla vision-based RAG pipeline with degraded input.
(c) The proposed \method{}, which introduces a causality-guided dual-path encoder to disentangle semantic and degradation factors.}

    \label{fig:model_overview}
\end{figure*}

\subsection{Preliminary}
\label{subsec:preliminary}

\noindent\textbf{Vision-based RAG.}  
Given a textual query \( q \) (e.g., a question or instruction) and a visual corpus \( \mathcal{V} = \{ X_i \}_{i=1}^{N} \),  
VisRAG retrieves the top-\(k\) most relevant document images and generates a response as follows:
{\small
\begin{equation}
\scalebox{0.95}{$
\underbrace{R}_{\text{top-}k\ \text{retrieved doc images}}
= \mathcal{R}\!\big(q,\; \mathcal{E}_r(\mathcal{V})\big),
\qquad
Y = \mathcal{G}\!\big(q,\; \mathcal{E}_g(R)\big),$}
\label{eq:visrag}
\end{equation}
where \( X_i \) denotes the \(i\)-th document image in the corpus,  
\( \mathcal{E}_r \) and \( \mathcal{E}_g \) represent the retrieval and generation encoders,  
and \( \mathcal{R}(\cdot) \) and \( \mathcal{G}(\cdot) \) denote the retrieval and generation modules, respectively.  
For notational simplicity, both encoders are hereafter referred to collectively as \( \mathcal{E}_\theta \).  

In real-world scenarios, document images inevitably suffer from visual degradations 
such as blur, noise, low light, and shadow,  
which distort semantics and cause distributional shifts in the embedding space.  
These degradations reduce retrieval accuracy and propagate errors into the generation stage.  
Even when correct document images are retrieved, degraded visual inputs can still mislead the generation process,  
resulting in hallucinated or semantically inconsistent responses.  
We attribute this issue to \textit{semantic--distortion entanglement},  
where semantic and degradation features become intertwined within the visual encoder representations.

\noindent\textbf{Causal Formulation of Degradation in VisRAG.}
We formalize how semantics and degradations jointly influence VisRAG outputs through a structural causal model (SCM).
Let $S$ denote the task-relevant semantic factors, and $D$ denote the degradation (nuisance) factors such as blur or shadow.
The observed document image $X$ is generated as
\begin{equation}
X = f(S, D, \varepsilon_X),
\end{equation}
where $f(\cdot)$ represents the image-formation process and $\varepsilon_X$ is an exogenous noise variable.
A pretrained vision encoder in VLMs $\mathcal{E}_\theta$ then maps $X$ into a latent representation:
\begin{equation}
Z = \mathcal{E}_\theta(X),
\end{equation}
which is further processed by the retrieval and generation modules:
\begin{equation}
(R, Y) = g(q, Z),
\end{equation}
where $g(\cdot)$ denotes the retrieval and generation mapping in the VisRAG pipeline.

Following the Independent Causal Mechanism principle~\cite{scholkopf2021toward}, 
we assume $S \perp\!\!\!\perp D$ and that all exogenous noise variables are mutually independent.
Here, $\perp\!\!\!\perp$ denotes statistical independence.
The overall causal structure can be summarized as
\begin{equation}
S \rightarrow X \leftarrow D, \qquad X \rightarrow Z, \qquad Z \rightarrow (R, Y),
\end{equation}
where the directed edges represent causal influences between variables. 
Specifically, $S$ and $D$ jointly determine the observed document image $X$; 
$X$ is then encoded into a latent representation $Z$ by $\mathcal{E}_\theta$; 
and $Z$ further drives the retrieval and generation outputs $(R, Y)$ in the VisRAG pipeline.
This structure clearly depicts how both semantic and degradation factors propagate through the encoding and reasoning stages.
Since $Z$ is a descendant of the collider $X$, conditioning on $Z$ can open a non-causal path $S \leftrightarrow D$ 
(\,$\leftrightarrow$ denotes an induced statistical association\,),
introducing residual dependence $S \not\!\perp\!\!\!\perp D \mid Z$.
Here, $\not\!\perp\!\!\!\perp$ denotes statistical dependence.
This entanglement leads the latent representation $Z$ to mix semantic and degradation information, 
resulting in corrupted retrieval and unstable generation.

To mitigate this issue, we propose to disentangle the two factors within the latent space. 
In particular, the representation encoded by $\mathcal{E}_\theta$ should preserve task-relevant semantics 
while isolating degradation cues into a separate subspace. 
Therefore, our objective is to learn a factorized representation:
\begin{equation}
Z = [Z_{\text{sem}}, Z_{\text{deg}}],
\end{equation}
where $Z_{\text{sem}}$ captures semantic content while $Z_{\text{deg}}$ encodes degradation information.
Ideally, the representation should satisfy:
\begin{equation}
Z_{\text{sem}} \not\!\perp\!\!\!\perp S, \quad 
Z_{\text{sem}} \perp\!\!\!\perp D\;.
\end{equation}
That is, the semantic component should depend only on the causal factors $S$ and remain independent of degradation factors $D$.
Under this condition, the prediction made by the model can be viewed as an approximation of the interventional distribution:
\begin{equation}
P(A \mid do(D=d_0)), \qquad A \in \{R, Y\},
\end{equation}
where the do-operator $do(\cdot)$~\cite{pearl2016causal} denotes intervention that removes the causal influence of $D$ (e.g., setting $D$ to a reference clean state $d_0$, or performing edge surgery to cut $D\!\to\!X$). 
$A$ as the retrieval and generation outputs $(R, Y)$. 
As shown in~\cref{fig:scm}, this intervention closes the non-causal path $D \!\to\! X \!\to\! Z \!\to\! A$ while preserving the causal route $S \!\to\! X \!\to\! Z \!\to\! A$. 
This causal formulation motivates the asymmetric encoder design and the disentanglement objectives introduced in Sec.~\ref{subsec:method}. For clarity, we omit the explicit variable $X$ in the following sections, since the encoder $\mathcal{E}_\theta$ implicitly maps the observed image $X$ into latent representations $(Z_{\text{sem}}, Z_{\text{deg}})$.

\subsection{\method{}}
\label{subsec:method}

\noindent\textbf{Overview.}
Building upon the causal analysis in \cref{subsec:preliminary},
\method{} enhances the standard VisRAG pipeline (\cref{fig:model_overview}(b))
with a causality-guided dual-path encoder (\cref{fig:model_overview}(c)).
The design explicitly follows the two information sources in the structural causal model:
a \textit{non-causal path} that captures degradation factors $D$,
and a \textit{causal path} that encodes task-relevant semantics $S$.
The non-causal path learns to identify and represent degradations through the
Non-Causal Distortion Modeling (NCDM) objective,
while the causal path leverages these learned degradation cues to disentangle
and purify semantic representations via the Causal Semantic Alignment (CSA) objective.
Both paths are jointly optimized in an end-to-end manner,
enabling structural intervention on the non-causal route and approximating the
interventional behavior $P(A \mid do(D=d_0))$.

\vspace{4pt}
\noindent\textbf{Non-Causal Path.}
To explicitly extract degradation-related information, we introduce a single
non-causal token $z_{nc}^{(0)}$ at the input layer. This token is propagated
through the network and updated at each layer. Let $\{x_1^{(l)}, \dots, x_T^{(l)}\}$ denote the patch tokens at layer $l$,
and let $z_{nc}^{(l)}$ denote the updated feature of the same non-causal token
after layer $l$. During attention computation, we enforce a directional constraint:
(i) the non-causal token is allowed to attend to all patch tokens, and
(ii) patch tokens are masked from attending to the non-causal token.
Under this design, the non-causal token is updated as
\begin{equation}
z_{nc}^{(l+1)}
= z_{nc}^{(l)}
+ \sum_{j=1}^{T} \alpha^{(l)}_{nc \leftarrow j} \, v_j^{(l)},
\label{eq:noncausal_forward_final_rewrite}
\end{equation}
where $\alpha^{(l)}_{nc \leftarrow j}$ denotes the attention weight from the
non-causal token (as query) to the $j$-th patch token, and $v_j^{(l)}$ is the
corresponding value projection.

This unidirectional design lets the non-causal token aggregate degradation cues
across spatial regions of the image while preventing these cues from flowing
back into the semantic tokens. After $L$ self-attention layers, we obtain the degradation
representation:
\begin{equation}
Z_{\text{deg}} = z_{nc}^{(L)}.
\end{equation}
However, structural separation alone does not guarantee that
$Z_{\text{deg}}$ truly captures degradation factors. Therefore, we further
introduce the NCDM objective to explicitly constrain this pathway to focus on
degradation modeling while maintaining its non-interference with semantic
representations.

\vspace{4pt}
\noindent\textbf{Non-Causal Distortion Modeling.}
The goal of the non-causal path is to encourage a degradation-aware latent subspace. This enables the causal (semantic) branch to
utilize such degradation embeddings as guidance for disentanglement
within the same forward pass.
To achieve this, we introduce a distortion-contrastive objective that enforces
degradation-aware discrimination in the latent space.
For an anchor image $X_a$ with degradation type $d_a$, we select a positive
sample $X_p$ sharing the same degradation ($d_p = d_a$) and a negative sample
$X_n$ with a different degradation ($d_n \neq d_a$), and obtain their
degradation embeddings
$Z^{a}_{\text{deg}}, Z^{p}_{\text{deg}}, Z^{n}_{\text{deg}}$ from the non-causal path.
The objective is formulated as
\begin{equation}
\mathcal{L}_{\text{NCDM}}
= \max\big(0,\;
\|Z^a_{\text{deg}} - Z^p_{\text{deg}}\|_2^2
- \|Z^a_{\text{deg}} - Z^n_{\text{deg}}\|_2^2
+ \delta \big),
\label{eq:ncdm_final}
\end{equation}
where $\delta$ is a margin parameter.
This contrastive formulation encourages $Z_{\text{deg}}$ to cluster samples with the
same degradation while separating those from different distortions, thereby
encouraging $Z_{\text{deg}}$ to encode degradation-consistent patterns that facilitate robustness, without enforcing degradation-type identifiability.

\vspace{4pt}
\noindent\textbf{Causal Path.}
In parallel to the non-causal branch, the causal branch focuses on semantic
aggregation through bidirectional attention among patch tokens. The non-causal
token is excluded from this attention, so that semantic encoding is not
contaminated by degradation-specific features.

Let $x_i^{(l)}$ denote the $i$-th patch token at layer $l$, where the subscript
$c$ implicitly marks the causal pathway. Given $T$ visual tokens, their update
rule is
\begin{equation}
x_i^{(l+1)} =
x_i^{(l)} +
\sum_{j=1}^{T} \alpha^{(l)}_{i \leftrightarrow j} \, v_j^{(l)},
\quad i = 1, \dots, T,
\label{eq:causal_path_final_rewrite}
\end{equation}
where $\alpha^{(l)}_{i \leftrightarrow j}$ is the attention weight between
patch tokens only, computed under a mask that removes the non-causal token from
both the key and value sets for the causal branch. This formulation is architecture-agnostic: for encoders with a global semantic
token~\cite{radford2021learning}, that token is included in the patch set and
attends to all patches; for pooling-based encoders~\cite{zhai2023sigmoid},
semantic aggregation arises from contextual interactions among patch tokens.

After $L$ layers, we obtain the semantic representation:
\begin{equation}
Z_{\text{sem}} = \mathrm{Agg}\big(x_1^{(L)}, \dots, x_T^{(L)}\big),
\end{equation}
where $\mathrm{Agg}(\cdot)$ denotes the semantic aggregation function used to aggregate patch tokens the underlying encoder architecture. This representation is expected to follow the causal route $S \!\to\! Z_{\text{sem}}$ and to remain invariant to degradations. Importantly, $Z_{\text{sem}}$ and $Z_{\text{deg}}$ are produced simultaneously within the same forward pass: $Z_{\text{deg}}$ provides a degradation-related constraint to guide disentanglement during training through the CSA, rather than forming a direct causal dependency on $Z_{\text{sem}}$.

\vspace{4pt}
\noindent\textbf{Causal Semantic Alignment.}
Despite structural separation, early-layer feature sharing may still cause
degradation leakage into the causal branch. To explicitly ensure that
$Z_{\text{sem}}$ captures degradation-invariant semantics, we propose the
CSA objective, which leverages $Z_{\text{deg}}$
as a causal regulator. For each clean/degraded image pair, we
extract $Z_{\text{sem}}^{\text{clean}}$ from the clean image’s causal path and
$(Z_{\text{sem}}^{\text{deg}}, Z_{\text{deg}}^{\text{deg}})$ from the degraded
image in the same forward pass. We define a joint semantic consistency and
independence loss:
\begin{equation}
\mathcal{L}_{\text{SIL}} = \frac{1}{T} \sum_{i=1}^{T}
\Big[(1 - \langle Z_{\text{sem},i}^{\text{deg}}, Z_{\text{sem},i}^{\text{clean}} \rangle)
+ \big|\langle Z_{\text{sem},i}^{\text{deg}}, Z_{\text{deg}}^{\text{deg}} \rangle\big|\Big],
\label{eq:sil_final}
\end{equation}
where $\langle \cdot, \cdot \rangle$ denotes cosine similarity and $T$ is the
number of semantic tokens. The first term aligns the degraded semantics with
the clean semantics, preserving the causal path
$S \!\to\! Z_{\text{sem}}$, while the second term enforces independence between
semantic and degradation embeddings, suppressing the non-causal dependency
$D \!\to\! Z_{\text{sem}}$. To maintain local structural consistency, we add a
fine-grained alignment term:
\begin{equation}
\mathcal{L}_{\text{FSAL}} =
\frac{1}{T} \sum_{i=1}^{T}
\big\| Z_{\text{sem},i}^{\text{deg}} - Z_{\text{sem},i}^{\text{clean}} \big\|_2^2.
\label{eq:fsal_final}
\end{equation}
The overall CSA objective is given by:
\begin{equation}
\mathcal{L}_{\text{CSA}} =
\mathcal{L}_{\text{SIL}} + \lambda_{\text{FSAL}} \mathcal{L}_{\text{FSAL}},
\label{eq:csa_final}
\end{equation}
where $\lambda_{\text{FSAL}}$ balances global and local alignment.
By jointly optimizing $\mathcal{L}_{\text{CSA}}$ and $\mathcal{L}_{\text{NCDM}}$,
the encoder learns to causally disentangle semantics from degradations within
a single forward pass, allowing $Z_{\text{sem}}$ to approximate the
interventional representation $P(Z_{\text{sem}} \mid do(D = d_0))$.

\vspace{4pt}
\vspace{4pt}
\noindent\textbf{Overall Objective.}
\method{} employs two vision encoders for retrieval and generation, both trained under the same causality-guided framework. 
For retrieval, the encoder is optimized end-to-end with a standard contrastive objective:
\begin{equation}
\mathcal{L}_{\text{Ret}} = -\log
\frac{\exp(\langle q, X^+ \rangle / \tau)}{
\exp(\langle q, X^+ \rangle / \tau) + \sum_{x^-} \exp(\langle q, X^- \rangle / \tau)},
\label{eq:retrieval_final}
\end{equation}
where $X^+$ and $\{X^-\}$ denote the positive and negative visual documents to query $q$, and $\tau$ is a temperature. 
The total retrieval loss combines contrastive learning with causality-guided objectives:
\begin{equation}
\mathcal{L}_{\text{Retrieval}} =
\mathcal{L}_{\text{Ret}}
+ \lambda_1 \mathcal{L}_{\text{CSA}}
+ \lambda_2 \mathcal{L}_{\text{NCDM}},
\label{eq:retrieval_total_final}
\end{equation}
where $\lambda_1$ and $\lambda_2$ balance causal alignment and degradation modeling. 
For generation, the language model remains frozen while the visual encoder is fine-tuned using only the causality-guided objectives:
\begin{equation}
\mathcal{L}_{\text{Generation}} =
 \mathcal{L}_{\text{CSA}}
+ \lambda_3 \mathcal{L}_{\text{NCDM}},
\label{eq:generation_total_final}
\end{equation}

This training allows the adapted encoder to maintain stable semantics under degraded conditions. 
Through these unified optimization strategies, \method{} enforces structural intervention to suppress the spurious route 
$D \!\to\! Z \!\to\! A$ while preserving the causal path $S \!\to\! Z_{\text{sem}} \!\to\! A$.

\vspace{4pt}
\noindent\textbf{Inference.}
At test time, we only require degradation-invariant semantics for retrieval and
generation.
Since the causal branch already produces $Z_{\text{sem}}$ that has been trained
to remove degradation information under the guidance of $Z_{\text{deg}}$, we discard
the non-causal branch and feed only $Z_{\text{sem}}$ to the downstream VisRAG
modules.
Thus, the inference-time computation and architecture remain compatible with
the standard VisRAG pipeline, while enjoying improved robustness to visual
degradations.

\subsection{Distortion-VisRAG Dataset}
We construct the Distortion-VisRAG (DVisRAG) dataset to evaluate the robustness of Vision-based RAG pipelines under degraded visual conditions.\footnote{More details on dataset statistics, collection procedures, and degradation generation methods are provided in the supplementary material.}
It covers seven major document VQA domains, including scientific papers, charts, slides, infographics, forms, handwritten notes, and reports, and contains a total of 367,608 question–document (Q–D) pairs.  
The dataset consists of two complementary parts: the Synthetic Degradation Dataset and the Real-World Degradation Dataset, corresponding to synthetic and real degradation scenarios, respectively.  
The synthetic subset includes 362,110 training samples and 3,607 testing samples, while the real subset contains 1,891 testing samples. 

\noindent\textbf{Synthetic Degradation Dataset.}  
This subset is built upon VisRAG~\cite{yu2024visrag}, encompassing all its original data sources, including ArXivQA~\cite{li2024multimodal}, ChartQA~\cite{masry2022chartqa}, PlotQA~\cite{methani2020plotqa}, InfoVQA~\cite{mathew2022infographicvqa}, MP-DocVQA~\cite{tito2023hierarchical}, SlideVQA~\cite{tanaka2023slidevqa} and Synthetic~\cite{yu2024visrag}.  
Following the degradation synthesis pipeline in UniRestore~\cite{chen2025unirestore}, we generate twelve common types of degradation (e.g., blur, noise, brightness variation, color saturation change, and resolution reduction), each at five severity levels. For each image in the dataset, we randomly sample one degradation type and one severity level to synthesize the degraded version. The training and testing splits are identical to those of VisRAG, and all question–answer pairs remain unchanged. Only the document images are degraded to ensure full comparability and consistent evaluation settings.

\noindent\textbf{Real Degradation Dataset.}  
This subset is designed to evaluate model generalization under real-world conditions.  
We randomly sample a portion of Q–D pairs from the ArXivQA and MP-DocVQA test sets of VisRAG, and additionally include document samples from RVL-CDIP~\cite{harley2015icdar}, yielding 1,891 non-overlapping test pairs.  
All documents are printed and photographed using a Sony RX100 VII camera under controlled capture settings, such as varying shutter speed, exposure compensation, and partial illumination occlusion, to create five real degradation types: blur, low light, low resolution, shadow, and paper damage.  
This dataset is used exclusively for testing and does not participate in training.
\section{Experiments}\label{sec:exp}
We conduct experiments to validate the effectiveness of \method{}. 
Due to space limitations, detailed implementation settings and more experimental results are provided in the supplementary material.

\subsection{Implementation Details.}
\method{} adopts the same backbone as VisRAG~\cite{yu2024visrag}, using MiniCPM-V~2.0~\cite{yu2024visrag} as the retriever and MiniCPM-V~2.6~\cite{minicpmv26} as the generator. Both components are initialized from pre-trained checkpoints and fine-tuned under the mixed-dataset setting, which combines the training splits of the VisRAG and DVisRAG datasets. Evaluation is conducted on three test sets: the VisRAG test split, the synthetic degradation subset, and the real degradation subset from DVisRAG. Following VisRAG~\cite{yu2024visrag}, we report MRR@10 (Mean Reciprocal Rank at 10) for retrieval ranking quality and Accuracy for generation performance, where the latter adopts a relaxed 5\% numerical tolerance to account for rounding and OCR errors.

\subsection{Baselines}
Following the evaluation protocol of VisRAG~\cite{yu2024visrag},
we evaluate \method{} across three stages: retrieval, generation, and end-to-end.
Baseline methods are categorized into two types:
(i) text-based pipelines that apply OCR~\cite{du2020pp} to extract textual content from document images and then perform retrieval or generation based on recognized text, and
(ii) vision-based pipelines that directly process raw images without textual conversion.
Text-based methods are marked with “(T)”, while unmarked ones are vision-based.

For retrieval, we consider multiple representative backbones, including
BM25 (T)~\cite{robertson1995okapi},
BGE-large (T)~\cite{bgeembedding},
NV-Embed-v2 (T)~\cite{lee2024nvembed},
SigLIP~\cite{zhai2023sigmoid},
and ColPali~\cite{faysse2024colpali}.
We also include MiniCPM-V2.0~\cite{yu2024visrag},
which serves as the retrieval backbone in VisRAG and is denoted as VisRAG-Ret.
For generation, we adopt several representative models, including
MiniCPM (T)~\cite{minicpmv2, yao2024minicpmv},
and GPT-4o~\cite{gpt4o},
We further include MiniCPM-V2.6~\cite{minicpmv26, yao2024minicpmv},
which serves as the generation backbone in VisRAG and is denoted as VisRAG-Gen.

To systematically analyze robustness and adaptation behavior,
we introduce three configurations as follows.
All models are fine-tuned following the official settings described in their respective papers,
ensuring consistent optimization protocols.

\noindent\textbf{(i) Vanilla Models.}
Pre-trained models are directly evaluated without any fine-tuning to assess their zero-shot capability.

\noindent\textbf{(ii) Fine-tuned on VisRAG Dataset.}
Following the official VisRAG training protocol~\cite{yu2024visrag}, models are fine-tuned on the VisRAG dataset using in-domain question–document supervision,
denoted as “\textit{-FV}”.

\noindent\textbf{(iii) Fine-tuned on Mixed Dataset.}
Models are fine-tuned jointly on the VisRAG dataset and our DVisRAG dataset, 
denoted as “\textit{-FM}”.

\begin{table}[t!]
    \centering
\caption{\textbf{Overall retrieval performance (MRR@10) on the VisRAG and the DVisRAG datasets.}
    ``Synthetic'' and ``Real'' denote results on the synthetic-degradation and real-degradation subsets of the DVisRAG dataset.}

    \label{tab:ret_vqa_avgonly}
    \scalebox{0.63}{
    \begin{tabular}{l c | c | cc}
        \toprule[1.3pt]
        \multirow{2}{*}{\textbf{Models}} & \multirow{2}{*}{\textbf{\# Para.}} 
        & \multirow{2}{*}{\textbf{VisRAG Dataset}} 
        & \multicolumn{2}{c}{\textbf{DVisRAG Dataset}} \\
        &  &  & \textbf{Synthetic} & \textbf{Real} \\
        \hline\hline
        BM25 (T)              & n.a         & 53.34 & 32.80 & 38.60 \\
        BGE-large (T)         & 335M        & 58.98 & 37.05 & 35.23 \\
        NV-Embed-v2 (T)       & 7.85B       & 72.41 & 47.68 & 49.44 \\
        MiniCPM-FV (T)            & 2.72B   & 74.94 & 48.54 & 52.09 \\
        MiniCPM-FM (T)           & 2.72B    & 73.17 & 49.96 & 51.72 \\
        SigLIP                & 883M        & 43.47 & 33.10 & 15.07 \\
        SigLIP-FV                 & 883M    & 71.52 & 57.51 & 28.50 \\
        SigLIP-FM                & 883M     & 68.84 & 59.77 & 31.89 \\
        ColPali-FV                & 2.97B   & 74.57 & 61.22 & 41.70 \\
        VisRAG-Ret             & 3.43B   & 77.57 & 65.96 & 56.47 \\
        VisRAG-Ret-FM            & 3.43B    & 78.39 & 68.69 & 58.25 \\
        VisRAG-Ret-FM (FARE)     & 3.43B    & \underline{78.42} & \underline{69.11} & \underline{59.39} \\        
        \method{}                & 3.43B    & \textbf{80.11} & \textbf{73.21} & \textbf{63.82} \\
        \bottomrule[1.3pt]
    \end{tabular}
    }
\end{table}

\begin{table}[t!]
    \centering
    \caption{\textbf{Overall generation performance (Accuracy) on the VisRAG and the DVisRAG datasets.}}
    \label{tab:gen_vqa_avgonly}
    \scalebox{0.6}{
    \begin{tabular}{l c | c | c c}
        \toprule[1.3pt]
        \multirow{2}{*}{\textbf{Models}} & \multirow{2}{*}{\textbf{Metric}} 
        & \multirow{2}{*}{\textbf{VisRAG Dataset}} 
        & \multicolumn{2}{c}{\textbf{DVisRAG Dataset}} \\
        &  &  & \textbf{Synthetic} & \textbf{Real} \\
        \hline\hline

        \multicolumn{1}{l}{\multirow{4}{*}{GPT-4o (T)}} 
            & top-1   & 44.05 & 26.26 & 27.61 \\
            & top-2   & 47.44 & 29.51 & 28.73 \\
            & top-3   & 47.03 & 30.63 & 29.44 \\
            & Oracle  & 53.06 & 37.90 & 41.54 \\
        \hdashline

        \multicolumn{1}{l}{\multirow{4}{*}{MiniCPM (T)}} 
            & top-1   & 28.44 & 18.58 & 18.21 \\
            & top-2   & 28.43 & 18.06 & 20.07 \\
            & top-3   & 27.79 & 18.18 & 18.80 \\
            & Oracle  & 31.92 & 24.23 & 25.22 \\
        \hdashline

        \multicolumn{1}{l}{\multirow{4}{*}{GPT-4o}} 
            & top-1   & 52.44 & 43.31 & 41.49 \\
            & top-2   & 53.69 & 45.29 & 44.08 \\
            & top-3   & 54.98 & 45.74 & 44.81 \\
            & Oracle  & \underline{63.50} & 54.52 & 58.61 \\
        \hdashline

        \multicolumn{1}{l}{\multirow{4}{*}{VisRAG-Gen}} 
            & top-1   & 51.51 & 42.16 & 44.70 \\
            & top-2   & 53.82 & 43.35 & 47.15 \\
            & top-3   & 54.11 & 44.53 & 47.14 \\
            & Oracle  & 63.36 & 51.52 & 62.68 \\
        \hdashline

        \multicolumn{1}{l}{\multirow{4}{*}{VisRAG-Gen-FM (PEFT)}} 
            & top-1   & 53.92 & 43.64 & 47.22 \\
            & top-2   & 55.44 & 44.51 & 48.82 \\
            & top-3   & 55.22 & 45.18 & 49.35 \\
            & Oracle  & 64.88 & 53.16 & \underline{63.52} \\
        \hdashline

        \multicolumn{1}{l}{\multirow{4}{*}{VisRAG-Gen-FM (FFT)}} 
            & top-1   & \underline{55.50} & 44.44 & 47.29 \\
            & top-2   & \underline{56.32} & 45.92 & 49.59 \\
            & top-3   & 55.68 & 46.17 & 49.03 \\
            & Oracle  & \underline{65.91} & 53.20 & 62.54 \\
        \hdashline

        \multicolumn{1}{l}{\multirow{4}{*}{VisRAG-Gen-FM (FARE)}} 
            & top-1   & 55.49 & \underline{45.68} & \underline{50.95} \\
            & top-2   & \underline{56.87} & \underline{47.75} & \underline{53.14} \\
            & top-3   & \underline{57.02} & \underline{48.18} & \underline{53.46} \\
            & Oracle  & \underline{66.13} & \underline{54.69} & \underline{64.46} \\
        \hdashline

        \multicolumn{1}{l}{\multirow{4}{*}{\method{}}} 
            & top-1   & \textbf{58.22} & \textbf{48.02} & \textbf{55.39} \\
            & top-2   & \textbf{60.98} & \textbf{53.18} & \textbf{57.91} \\
            & top-3   & \textbf{61.99} & \textbf{54.01} & \textbf{59.44} \\
            & Oracle  & \textbf{67.33} & \textbf{57.87} & \textbf{69.03} \\
        \bottomrule[1.3pt]
    \end{tabular}}
\end{table}

\begin{table}[t!]
    \centering
    \caption{\textbf{End-to-end retrieval–generation performance on VisRAG and DVisRAG datasets.}}
    \label{tab:end2end_performance}
    \scalebox{0.58}{
    \begin{tabular}{l | ccc | ccc}
    \toprule[1.3pt]
    \multirow{3}{*}{\textbf{Methods}} 
      & \multicolumn{3}{c|}{\textbf{Retrieval (MRR@10)}} 
      & \multicolumn{3}{c}{\textbf{Generation (Top-1)}} \\ 
    \cmidrule(lr){2-4} \cmidrule(lr){5-7}
      & \textbf{VisRAG Dataset} 
      & \textbf{Synthetic} 
      & \textbf{Real}
      & \textbf{VisRAG Dataset} 
      & \textbf{Synthetic} 
      & \textbf{Real} \\
    \midrule
    VisRAG 
      & 77.57 
      & 65.96 
      & 56.47 
      & 50.40 
      & 41.96 
      & 42.99 \\
    VisRAG-FT 
      & \underline{78.42} 
      & \underline{69.11} 
      & \underline{59.39} 
      & \underline{54.84} 
      & \underline{44.96} 
      & \underline{48.27} \\
    Two-Stage 
      & 77.78 
      & 66.49 
      & 53.59 
      & 50.56 
      & 42.25 
      & 40.42 \\
    \method{} 
      & \textbf{80.11} 
      & \textbf{73.21} 
      & \textbf{63.82} 
      & \textbf{58.22} 
      & \textbf{48.02} 
      & \textbf{55.39} \\
    \bottomrule[1.3pt]
    \end{tabular}}
\end{table}

\subsection{Overall Results and Analysis}
\noindent\textbf{Retrieval Performance.}
As shown in~\cref{tab:ret_vqa_avgonly}, \method{} achieves the best retrieval performance across all datasets.
Compared with the original VisRAG-Ret, \method{} improves retrieval accuracy by 2.54\% on clean data and by 7.25\% and 7.35\% under synthetic and real degradations.
We also compare against VisRAG-\textit{FARE}, which applies adversarial robustness training to VisRAG-Ret.
Even under this stronger baseline, \method{} achieves further gains of +1.69\%, +4.10\%, and +4.43\% on the clean, synthetic, and real subsets, respectively. 
Three observations emerge under our setting.
First, vision-based retrieval is inherently more stable under degradations, while OCR-dependent pipelines suffer from noise, blur, and illumination artifacts.
Second, mixed-dataset fine-tuning (\textit{-FM}) consistently improves degraded-domain performance, though its effect on clean accuracy varies across architectures.
Third, adversarial robustness training yields limited improvement under the complex degradations scenarios in DVisRAG dataset.
In contrast, \method{} explicitly disentangles semantic and degradation factors, leading to robustness that generalizes consistently across all visual conditions.

\vspace{3pt}
\noindent\textbf{Generation Performance.}
We evaluate various generation models using the retrieval results obtained from \method{}.  
Note that in the original VisRAG~\cite{yu2024visrag}, only the retriever was fine-tuned, while the generation module (VisRAG-Gen) remained frozen.  
To further investigate how generator adaptation affects robustness, we fine-tune VisRAG-Gen under three strategies:  
Full Finetuning (denoted as "\textit{-FFT}"),  
PEFT~\cite{hu2022lora} (denoted as "\textit{-PEFT}"),  
and adversarial robustness training following FARE~\cite{schlarmann2024robust} (denoted as "\textit{-FARE}").  

We report results using the top-1, top-2, and top-3 retrieved documents, as well as under the Oracle setting, where the model is given access only to the ground-truth positive document.  
As shown in~\cref{tab:gen_vqa_avgonly}, \method{} consistently outperforms all existing methods across different settings, achieving stable improvements on both synthetic and real-world degraded datasets.  
Specifically, \method{} improves over VisRAG-Gen by 6.35\% under the Oracle setting and surpasses GPT-4o by 10.42\%. Among the fine-tuning strategies, FARE achieves better robustness than FFT and PEFT due to its additional feature-space alignment constraint, which helps the model resist local perturbations.  
However, its improvement remains limited since such alignment does not explicitly disentangle semantic and degradation representations.  
In contrast, \method{} leverages degradation features extracted from the non-causal path as guidance to explicitly separate these factors during training, achieving stronger semantic stability and degradation invariance across both clean and corrupted inputs.

\vspace{3pt}
\noindent\textbf{End-to-End Performance.}
We further evaluate the complete retrieval–generation pipeline to assess the end-to-end robustness of \method{} compared with VisRAG-based configurations.  
Since VisRAG~\cite{yu2024visrag} and \method{} share identical retrieval and generation backbones, their differences lie solely in training and adaptation strategies.  
We include the following variants for comparison:  
(i) the vanilla VisRAG (denoted as \textit{VisRAG});  
(ii) the best-performing VisRAG fintuning configuration combining VisRAG-Ret-FM (FARE) and VisRAG-Gen-FM (FARE) (denoted as \textit{VisRAG-FT});  
and (iii) a two-stage enhancement strategy, where degraded images are first restored using image restoration method~\cite{potlapalli2023promptir} before being fed into the vanilla VisRAG pipeline (denoted as \textit{Two-Stage}).  

As shown in~\cref{tab:end2end_performance}, \method{} outperforms all baselines under degraded conditions while maintaining comparable accuracy to the vanilla VisRAG in clean settings.
On real-world degraded datasets, \method{} achieves an average improvement of 7.35\% in the retrieval stage and further raises the end-to-end accuracy by 12.4\%, indicating that the benefits of semantic–degradation disentanglement effectively propagate through the entire pipeline.
In contrast, the two-stage enhancement strategy, though conceptually intuitive, offers limited gains since the restoration step may distort clean images and does not ensure downstream robustness under degraded conditions.

\begin{table}[t!]
    \centering
    \caption{\textbf{Ablation on different configurations of \method{} on DVisRAG dataset.}}
    \label{tab:abla_modules}
    \scalebox{0.62}{
    \begin{tabular}{l | cc | cc}
    \toprule[1.3pt]
    \multirow{2}{*}{\textbf{Configurations}} 
    & \multicolumn{2}{c|}{\textbf{Retrieval (MRR@10)}} 
    & \multicolumn{2}{c}{\textbf{Generation (Top-1)}} \\ 
    \cmidrule(lr){2-3} \cmidrule(lr){4-5}
    & \textbf{Synthetic} & \textbf{Real} 
    & \textbf{Synthetic} & \textbf{Real} \\
    \midrule
    Baseline                                    
        & 65.96 & 56.47 & 41.96 & 42.99 \\
    \method{} w/o U                             
        & 69.12 & 60.28 & 45.34 & 49.54 \\
    \method{} w/o $\mathcal{L}_{\text{NCDM}}$   
        & 69.20 & 61.94 & 47.21 & 51.79 \\
    \method{} w/o $\mathcal{L}_{\text{CSA}}$    
        & 67.48 & 58.24 & 44.96 & 45.72 \\
    \method{} w/o $\mathcal{L}_{\text{NCDM}}$ \& $\mathcal{L}_{\text{CSA}}$ 
        & 66.34 & 56.94 & 42.94 & 43.80 \\
    \method{} 
        & \textbf{73.21} & \textbf{63.82} & \textbf{48.02} & \textbf{55.39} \\
    \bottomrule[1.3pt]
    \end{tabular}}
\end{table}

\begin{figure}[t!]
  \centering
  \setlength{\tabcolsep}{2pt}
  \footnotesize

  \begin{tabular}{ccc}
    \includegraphics[width=0.15\textwidth]{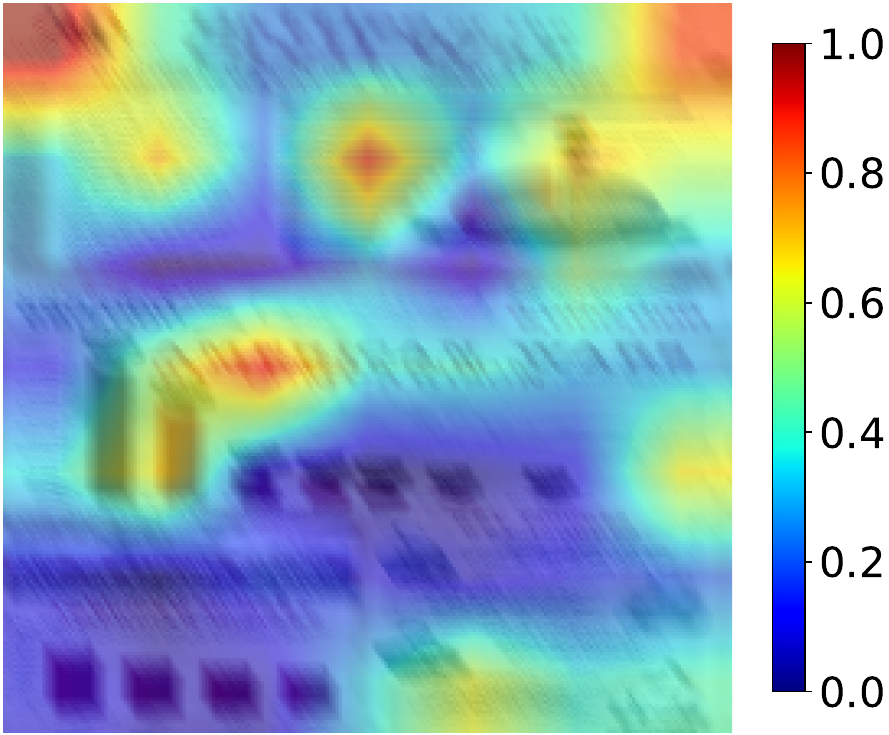} &
    \includegraphics[width=0.15\textwidth]{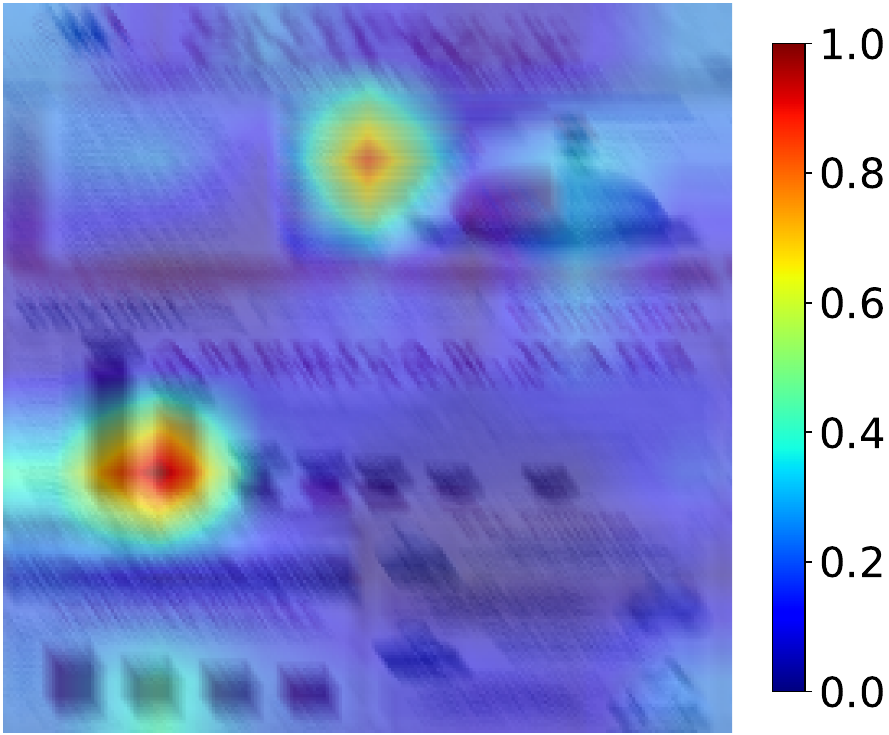} &
    \includegraphics[width=0.12\textwidth]{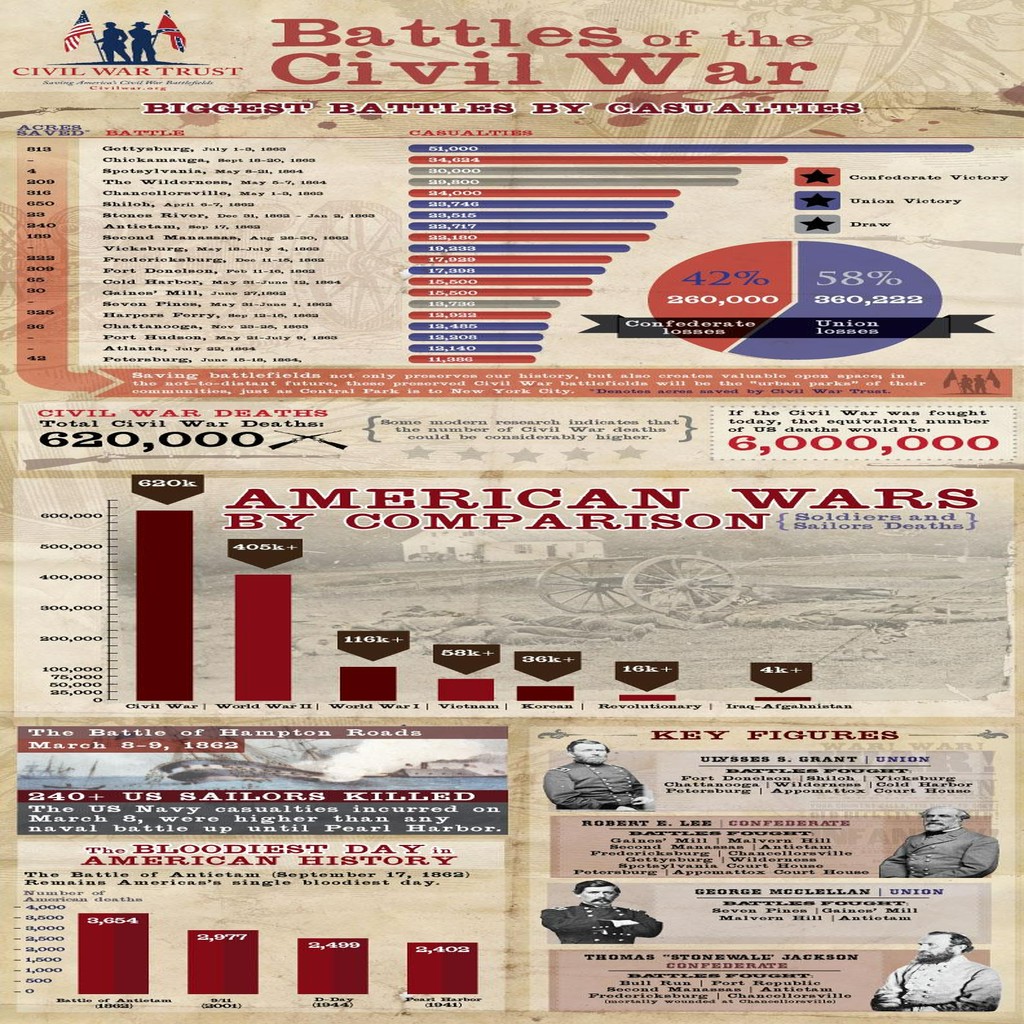} \\
    (a) & (b) & (c)
  \end{tabular}

  \vspace{6pt}

  \begin{tabular}{cc}
    \includegraphics[width=0.18\textwidth]{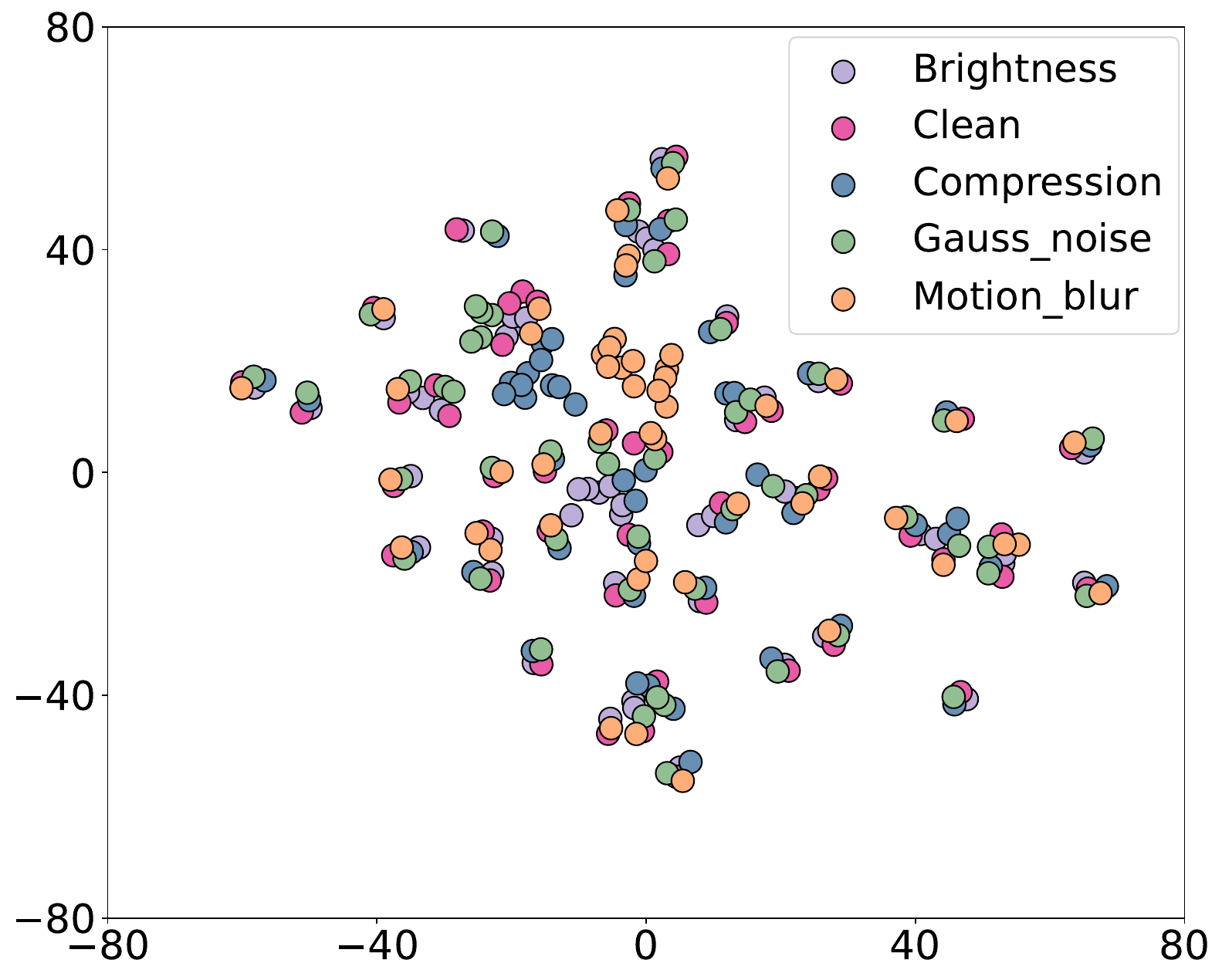} &
    \includegraphics[width=0.18\textwidth]{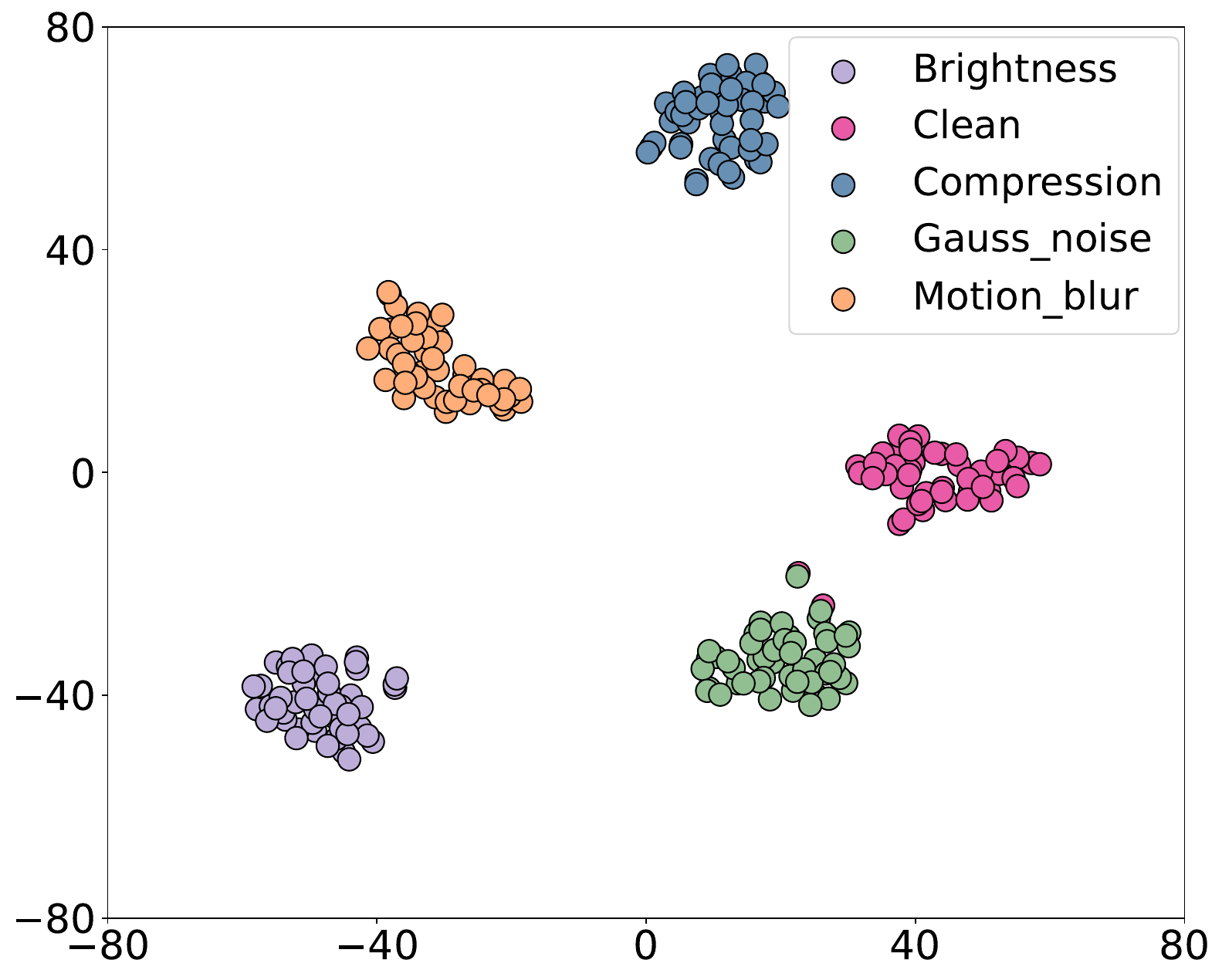} \\
    (d) & (e)
  \end{tabular}
\caption{
\textbf{Comparison of token representations under degradations:}
Attention visualizations of (a) VisRAG and (b) \method{}.  
(c) Clean version corresponding to (a) and (b).  
t-SNE visualization of $Z_{\text{deg}}$ from (d) \method{} w/o $\mathcal{L}_{\text{NCDM}}$ \& $\mathcal{L}_{\text{CSA}}$ and (e) \method{}.
}

  \label{fig:ncp_abla}
\end{figure}

\subsection{Ablation Study}
To analyze the contribution of each component, we train all variants on the mixed dataset and evaluate them on both the VisRAG and DVisRAG test sets.

\noindent\textbf{Effectiveness of Proposed Modules.}
We design six configurations to analyze the contribution of each component:
(i) Baseline: the original VisRAG framework;
(ii) \method{} w/o U: replacing the unidirectional Non-Causal Path with a bidirectional connection. This setting is equivalent to adding a Non-Causal Token into the VisRAG architecture but training it jointly with the two objectives $\mathcal{L}_{\text{NCDM}}$ and $\mathcal{L}_{\text{CSA}}$, without enforcing directional separation;
(iii) \method{} w/o $\mathcal{L}_{\text{NCDM}}$: removing the non-causal degradation modeling objective;
(iv) \method{} w/o $\mathcal{L}_{\text{CSA}}$: removing the causal semantic alignment objective;
(v) \method{} w/o $\mathcal{L}_{\text{NCDM}}$ \& $\mathcal{L}_{\text{CSA}}$: removing both loss terms simultaneously;
(vi) \method{}: the full model with all proposed modules. 
As shown in~\cref{tab:abla_modules}, all components contribute to improved robustness and generalization. The unidirectional attention constraint is essential for preventing semantic–degradation entanglement and preserving a clear causal separation between the two pathways, as evidenced by the comparison between (ii) and (vi). The comparison between (v) and (vi) further shows that adding a Non-Causal Path alone is insufficient; without the two proposed objectives, it fails to learn meaningful degradation features and yields only limited gains. Overall, the results demonstrate that each module in \method{} is necessary.

\noindent\textbf{Investigation of Learned token representations.}  
To analyze how degradations affect semantic encoding, we conduct two complementary visualization studies.
First, we sample a degraded image from DVisRAG and use the text query \texttt{“Bar Chart”}.
We compute the similarity between the text embedding and the mean patch-token features, then project the similarity map back onto the image.
As shown in~\cref{fig:ncp_abla}(a)(b), \method{} focuses more consistently on semantically relevant regions, whereas vanilla VisRAG is easily disrupted by degradations and tends to highlight irrelevant areas.
This indicates that \method{} learns semantic representations that are significantly more degradation-invariant.

Next, we sample 50 image–question–answer triplets and apply five types of synthetic degradations to each image. We then compare the degradation representations using the $Z_{\text{deg}}$ features from \method{} and from the variant \method{} w/o $\mathcal{L}_{\text{NCDM}}$ \& $\mathcal{L}_{\text{CSA}}$.
As shown in~\cref{fig:ncp_abla}(c)(d), the variant without these objectives exhibits poor separability among degradation types, whereas \method{} produces clear and compact clusters. This demonstrates that the combined effect of NCDM and CSA encourages degradation-consistent structure in the latent space. 

\section{Conclusion}
We presented \method{}, a VisRAG-oriented causality-guided dual-path framework that mitigates retrieval–generation error propagation under degradations. Through structural design and targeted objectives, \method{} improves retrieval, generation, and end-to-end performance under degradations while preserving clean-data accuracy. These gains come with no additional inference cost. We also introduce the \dataset{} dataset, a comprehensive benchmark for evaluating multimodal RAG models under degraded visual conditions.

\clearpage

{\small
\bibliographystyle{ieee_fullname}
\bibliography{egbib}
}

\end{document}